\documentclass[10pt,letterpaper]{article}

\usepackage{cogsci}
\usepackage{graphicx}
\usepackage{booktabs}
\usepackage{amsmath}
\usepackage{amssymb}
\usepackage{caption}

\cogscifinalcopy %

\usepackage[
    backend=biber,
    style=apa,
    natbib=true,
    doi=false,
    isbn=false,
    url=false,
]{biblatex}
\usepackage{pslatex}
\usepackage{float} %

\title{ Are words equally surprising in audio and audio-visual comprehension? }
 
\author{{\large \textbf{Pranava Madhyastha}} (pranava.madhyastha@city.ac.uk) \\
  Department of Computer Science,  City, University of London
  \AND {\large \bf Ye Zhang (y.zhang.16@ucl.ac.uk) }\\
  Division of Psychology and Language Sciences, University College London
  \AND {\large \bf Gabriella Vigliocco (g.vigliocco@ucl.ac.uk)} \\ 
  Division of Psychology and Language Sciences, University College London
  }

\begin{document}

\maketitle

\begin{abstract}
We report a controlled study investigating the effect of visual information (i.e., seeing the speaker) on spoken language comprehension. We compare the ERP signature (N400) associated with each word in audio-only and audio-visual presentations of the same verbal stimuli. We assess the extent to which surprisal measures (which quantify the predictability of words in their lexical context) are generated on the basis of different types of language models (specifically $n$-gram and Transformer models) predict N400 responses for each word. Our results indicate that cognitive effort differs significantly between multimodal and unimodal settings. 
In addition, our findings suggest that while Transformer-based models, which have access to a larger lexical context, provide a better fit in the audio-only setting, 2-gram language models are more effective in the multimodal setting. This highlights the significant impact of local lexical context on cognitive processing in a multimodal environment.

\end{abstract}

\section{Introduction}
A significant amount of research in language comprehension has been dedicated to examining how humans interpret written or spoken language. These studies have mainly focused on analyzing the verbal form of language \citep{fenk1980konstanz,asr-demberg:2015uniform-surprisal,jaeger2006speakers,hale2001probabilistic}. This approach involves building an understanding of the text or speech one word at a time, with some words being more difficult to process than others.
Expectation-based theories of sentence processing \citep{levy2008expectation} propose that the difficulty in processing a sentence is driven by the predictability of upcoming lexical material in context. Surprisal, an information-theoretic measure of predictability, is computationally operationalised using language models \citep{jaeger2006speakers,meister2021revisiting,aurnhammer2019evaluating,michaelov2020well}. Language models (LMs) calculate the probability of a word given its context, which is then used to calculate surprisal.  Surprisal has been supported by behavioural and neural measures of processing difficulty \citep{jaeger2006speakers,frank2015erp,aurnhammer2019evaluating}.  

However, a large body of previous work in language comprehension does not consider the visual contextual cues available in face-to-face communication. 
Language has evolved, is learnt and is most often used in face-to-face contexts in which comprehenders have access to a multitude of visual cues, such as hand gestures, body movements and mouth movements that contribute to language processing \citep{holler2019multimodal}. In this paper, we follow this line of research and examine how multiple modalities of information impact language comprehension. We present a controlled study comparing the comprehension of language-related stimuli in both audio-only and audio-visual conditions and analyse changes in ERP signals.

\subsection{N400, Language Models and Surprisal}

The N400 is an event-related potential (ERP) component peaking negatively at $\approx$400ms at the central parietal areas that are observed in the brain during language \textit{processing tasks},  measured using electroencephalography (EEG). The N400 is larger in response to semantically incongruent or unexpected words compared to congruent or expected words \citep{kutas1980reading,kutas2011thirty,kuperberg2016separate,michaelov2020well}. This indicates that the N400 is related to semantic processing, and the N400 effect has been interpreted as reflecting the brain's automatic evaluation of incoming linguistic information for semantic coherence Typically, when an upcoming word is semantically consistent with the context, it leads to a smaller N400 amplitude compared to when it is not. 

It has been reported in reading-related tasks that words with higher surprisal, thus less predictable\footnote{predictability is estimated using language models} and more difficult to process, elicit more negative N400 \citep{aurnhammer2019evaluating, delogu2017teasing, frank2015erp, merkx2020human, michaelov2020well, michaelov2021different, michaelov2022more}. Previous research has demonstrated the robustness of the N400 effect, and surprisal has been shown to predict N400 for various experimental tasks, including cloze-style tasks and semantic relatedness, among others \citep{michaelov2020well}. Recent work observe that surprisal estimates computed using some types of language models may be better predictors than other types. For example, \citet{frank2015erp} %
find that $n$-gram based language models with larger window sizes (4-grams) were best at explaining variance. More recent works have investigated Transformer based language models \citep{vaswani2017attention} and show that Transformer based models may be better predictors of suprisal than other language models. \citet{michaelov2021different} compared surprisal obtained from GPT-2 \citep{radford2019language} (a Transformer based language model trained over large web-based corpora), Recurrent Neural Network (RNN) based language model \citep{elman1990finding} and manual cloze probability in predicting N400 in cloze tasks, where the target words are manipulated to have different cloze probabilities. 
The authors discovered that all three measures showed a significant association with N400, but surprisal estimates generated from GPT-2 explained the largest amount of variance. Merkx and Frank (2020) conducted a study in which they trained language models with Transformer and RNN-based models in a controlled setting using similar corpora. Under these controlled conditions, the study found that surprisal estimates generated from Transformer-based models, overall, provided a better fit to the EEG data. The increased performance has been hypothesized to be primarily due to the access to a larger lexical context in Transformer-based language models, which helps the model capture longer-range dependencies.

Overall, most recent works have shown that surprisal estimates from Transformer based models correlate better with N400 based estimates of cognitive effort. We note that
the majority of the N400 and surprisal correlations were found in reading based tasks. Some recent works have shown that surprisal also predicts N400 based cognitive effort in audio \citep{brennan2019hierarchical} and audio-visual tasks \citep{zhang2021more, zhang2021electrophysiological} contexts. 
However, it remains unclear whether Transformer-based models such as GPT-2 are better predictors in audio or audio-visual settings where multiple sources of information are available.

\subsection{Multimodality, Surprisal and N400}

 Language learning and use is fundamentally face-to-face, involving  information from multiple sources (or modalities) such as gestures, facial expression, mouth movements and prosody, in addition to the lexical content of speech %
 \citep{zhang2021more, zhang2021electrophysiological}.  These modalities provide additional context and meaning, making communication more effective \citep{grzyb2022communicative,ankener2018influence,zhang2021more, zhang2021electrophysiological}. Recent studies have shown that multiple modalities, such as prosody (the rhythm, stress, and intonation of speech) and gestures, play a key role in shaping language use during face-to-face interactions and in general language use. 
 
Crucially, multimodal information, such as prosody and gesture, also modulates N400. For example, prosodic stress has been shown to mark the information structure, with new information more likely to carry prosodic stress than lexical information \citep{cruttenden2006accenting, aylett2004smooth}. Violations of such patterns elicit larger N400 \citep{magne2005line,wang2011influence, dimitrova2012less, baumann2012accentuation}, indicating that prosodic information is taken into account in semantic processing. Crucially also visual signals such as iconic gestures (hand movements imagistically related to the content of speech, e.g., "drawing" - imitate holding a pen and moving around) have been shown to affect the N400. Iconic gestures that mismatch the speech elicit larger N400 \citep{wu2005meaningful, kelly2004neural, ozyurek2007line, holle2007role}, indicating enhanced semantic processing difficulty. 
 
\citet{zhang2021more} further investigated how multimodal information modulates N400 in the naturalistic context where different cues co-occur. The authors in this study present participants with videos where a speaker produces short passages with naturally occurring prosody, gestures and mouth movements. They then quantified the correlation between
the lexical predictability (using $2$-gram surprisal estimates), 
prosody (using mean F0, capturing the pitch of the word), 
gestures (annotated as meaningful, e.g. ``drinking" - imitate holding a cup to drink, or beats, the rhythmic hand movements that are not directly meaningful), and informativeness of mouth movements. %
This study shows that ERP between 300-600ms is indeed sensitive to surprisal, extending the previous N400-surprisal effect to audio-visual modality. However, they also found that the effect of surprisal on N400 is modulated by multimodal information, as pitch prosody, meaningful gestures and informative mouth movements and their combinations reduce the N400, especially for higher surprisal words, indexing easier comprehension than predicted by surprisal alone. \citet{zhang2021electrophysiological} further report similar patterns in highly proficient non-native English comprehenders. 
These findings indicate that surprisal may not fully capture comprehension in the multimodal context, as the surprisal effect is modulated by multimodal information. However, both these studies
only use a $2$-gram based language model to compute surprisal estimates. It is unclear whether other models such as Transformers (which have access to a larger window of context) would allow for a better fit for N400 in the audio and audio-visual context. %
 \citet{ankener2018influence} presented evidence showing that visual information can impact lexical expectations in reading and listening experiments. They determined the index of cognitive activity by examining the impact of visual uncertainty on word surprisal and cognitive effort. These experiments focused on presenting additional visual stimuli that matched the words in the sentences. These findings suggest that in a controlled environment where visual stimuli are carefully provided, they have a significant effect on cognitive processing. This indicates the importance of taking into account additional information channels besides lexical content to accurately predict cognitive effort.
\subsection{The Present Study}

We report a controlled study to investigate the effects of visual signals (seeing the speaker) on language comprehension.  We compare the effects of audio-only and audio-visual settings using the same language stimuli and analyze the changes in ERP signals. We then evaluate the effectiveness of surprisal estimates, using different language models with varying lexical context windows, in explaining cognitive effort in both unimodal (audio-only) and multimodal (audio-visual) conditions.

Our study extends recent observations that indicate that other modalities of information significantly contribute towards the cognitive effort of language processing \citep[\emph{inter alia}]{zhang2021electrophysiological,zhang2021more,ankener2018influence}. We provide a comparison of EEG responses to the same lexical context but presented in a unimodal or multimodal manner. Crucially,  our analysis of language model surprisal estimates assesses whether language models with different architectures and degrees of complexities provide equally good fit across unimodal and multimodal contexts. 
We first present our methodology followed by the results and finally discuss the salient observations in the following sections.

\section{Methods}

\subsection{Electrophysiological Data}

\subsubsection{Participants}

We collected experimental data from two cohorts: a) 27 participants  %
in the audio-only condition and %
b) 31 participants in the audio-visual condition. All participants were native English speakers with normal hearing, vision, and no known neurological disorder\footnote{The study was approved by the university ethics committee. Participants gave written consent and were paid £7.5/h for their participation.}. 

\subsubsection{Materials}

103 naturalistic passages were randomly selected from the British National Corpus (BNC) and were evaluated by native English speakers to be semantically and grammatically coherent. They were recorded by a native English-speaking actress with natural prosody and facial expressions. The final corpus of experimental stimuli has a mean duration of 8.50 seconds and  an average word count of 23. The onset and offset of each word were automatically detected using a word-phoneme aligner based on a Hidden Markov Model \citep{rapp1995automatic} and was further manually verified (mean=440ms, SD=376ms). Participants watched the videos in the audio-visual setting and listened to the soundtrack of the videos in the audio-only setting.

\subsubsection{Procedures}

Participants were seated approximately 1 meter away from a computer and wore earphones during the experiment. After three practice trials, they were presented with audio stimuli in the first experiment and audio-visual stimuli in the second experiment. To ensure comparability between the two experiments, participants in the first experiment also viewed a static snapshot of the same actress taken from the video, to control for the presence of visual input. Each trial was separated by a 2000ms interval, and 35 clips were followed by attention checks to ensure participants were paying attention to the stimuli. Participants were instructed to carefully listen to or watch the stimuli and answer as quickly and accurately as possible.

The EEG data was collected for both the audio-only and audio-visual conditions using the same 32-channel Biosemi system with CMS and DRL as ground reference. Two external electrodes were attached to the left and right mastoid as an offline reference, and two external electrodes captured horizontal and vertical eye movements. Participants were instructed to avoid moving, keep their facial muscles relaxed, and reduce blinking, if possible. The electrode offsets were maintained between $\pm$ 25mV. The recording was conducted in a shielded room with a temperature of 18 °C. The EEG session lasted approximately 60 minutes.

\subsubsection{EEG Preprocessing}

The data was pre-processed with EEGLAB (\cite{delorme2004eeglab}, v.14.1.1) and ERPLAB (\cite{lopez2014erplab}, v.7.0.0) running under MATLAB 2019a. All electrodes were included. While N400 has a central-parietal distribution, the scalp distribution of audio and audiovisual speech can be more frontal and may be different from one another due to the modality differences \citep{kutas2011thirty}. Therefore, instead of focusing on a predefined region of interest (ROI), we included all electrodes \citep{zhang2021more, zhang2021electrophysiological}, categorized them into ROIs and added them in the statistical model (as in \cite{michaelov2021different}, see Statistical Analysis section below for more description). EEG files were referenced to mastoids, down-sampled to 512Hz, separated into -100 to 1200ms epochs time-locked to word onset and filtered with a 0.05-100Hz band-pass filter. Artefacts (e.g., eye movements and muscle noise) were first corrected with ICA. The remaining artefacts were rejected using a moving window peak-to-peak analysis and step-like artefact analysis. Due to likely overlap between any baseline period (-100 to 0ms) and the EEG signal elicited by the previous word, we did not perform baseline correction, but instead extracted the mean EEG amplitude in this time interval and later used it as a control variable in the statistical analysis \citep[see also][]{frank2015erp}. Following previous work \citep{frank2015erp, zhang2021more}, we take the mean ERP amplitude between 300-500ms as the N400 signal.

\subsection{Computing Surprisal}
Surprisal theory \citep{hale2001probabilistic,levy2008expectation,boston2008parsing} is rooted in information theoretic principles \citep{shannon1948mathematical} by utilising entropy, a core concept in information theory, to assess the predictability of events and the level of surprise they generate. The theory examines the connection between predictability and the processing of lexical information in the human brain. In this framework, lexical units carry information which is conveyed through a probabilistic measure. The level of predictability of these units influences how the brain processes and evaluates them. When predictability is low, it results in higher levels of surprise and requires more cognitive resources for processing. The exact amount of information conveyed by a unit is hence quantified as its \textit{surprisal}.

Formally, consider a linguistic signal $\mathbf{l}$ made of units: $\{l_1, \cdots, l_n\}$ (where the units could be words, phonemes, etc.); surprisal is then defined as:
\begin{equation}
s(l_t) = -\log p(l_t \mid l_1, \cdots, l_{t-1})
\label{logprob}
\end{equation}
which represents the negative log-probability of a unit $(l_t)$ given its preceding context $(l_1, \cdots, l_{t-1})$, where $t$ indicates the sequence time-steps. Surprisal theory asserts that the effort needed to process a linguistic unit is directly proportional to its unexpectedness in its context, which is measured by its surprisal.  Formally, for a linguistic unit $(l_t)$, the processing effort is linearly proportional to its surprisal:
\[
\text{effort}(l_t) \propto s(l_t)
\]
\begin{figure}[!htbp]
  \begin{center}
  \includegraphics[width=\linewidth,height=9cm]{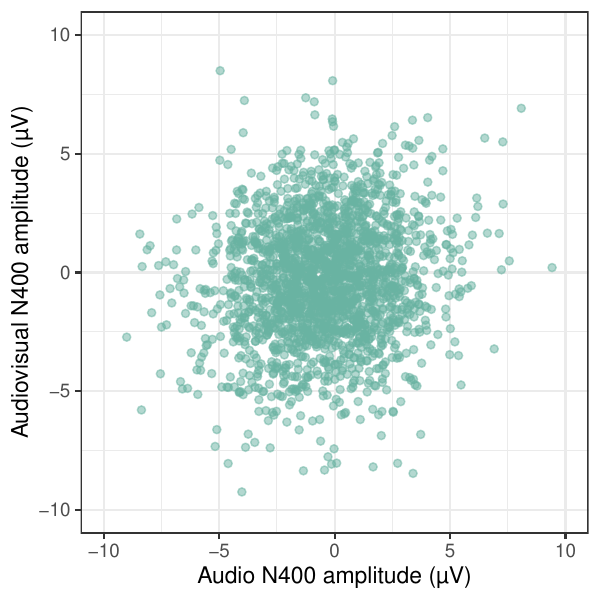}
  \caption{Scatter plot of N400 signal across audio-only and audio-visual modalities (Pearson correlation $r=0.11$). %
  The plot showcases that there is a weak correlation between the two settings for the same lexical input.} 
  \label{fig1}
  \end{center}
  \label{mainfig}
\end{figure}
As we don't have direct access to the true conditional probabilities of observing linguistic units given their context, we use language models to estimate them instead. We obtain surprisal estimates using log-probabilities (see Equation~\ref{logprob} above) through classical $n$-gram-based language models and more recent Transformer-based models.

For $n$-gram models, we cover an entire spectrum of $n$-gram models and construct  \{2,3,4,5,6\}-gram models using modified Kneser-Ney Smoothing \citep{ney1994structuring}\footnote{Following \citet{meister2021revisiting}, we use Wiki-text 103 as the corpus for estimating the $n$-gram probabilities.}. All probability estimates are computed at the word level. For Transformer-based models, we use GPT-2 and BERT \footnote{We use openly available pre-trained models from hugginface library \citep{wolf2020transformers}.}, and all probability estimates are also computed at the word level. We note that BERT is trained for a cloze-style task and hence the probabilities from this model are considered as pseudo surprisal estimates.
\begin{table*}[h]
    \begin{minipage}{.5\textwidth}
        \centering
        \caption*{Model comparisons in audio-only setting} 
        \label{tab:first}
        \resizebox{0.8\linewidth}{!}{
\begin{tabular}{lrrlrr} 
\toprule
 & Model & $\tilde{\chi}^2$ & Df & $p$ & $\tilde{\Delta}AIC$ \\
\midrule
$2$-gram & surprisal+ROI & 17.76 & 1.00 & $<.001$ & 15.76\\
 & $\Delta$surprisal$\times$ROI & 10.03 & 5.00 & 0.07 & 15.80\\
\midrule
$3$-gram & surprisal+ROI & 21.46 & 1.00 & $<.001$ & 19.46\\
& $\Delta$surprisal$\times$ROI & 20.05 & 5.00 & $<.001$ & 29.51\\
\midrule
$4$-gram & surprisal+ROI & 34.64 & 1.00 & $<.001$ & 32.64\\
& $\Delta$surprisal$\times$ROI & 20.74 & 5.00 & $<.001$ & 43.37\\
\midrule
$5$-gram & surprisal+ROI & 33.02 & 1.00 & $<.001$ & 31.02\\
 & $\Delta$surprisal$\times$ROI & 20.60 & 5.00 & $<.001$ & 41.62\\
\midrule
$6$-gram & surprisal+ROI & 33.96 & 1.00 & $<.001$ & 31.96\\
 & $\Delta$surprisal$\times$ROI & 21.09 & 5.00 & $<.001$ & 43.05\\
\midrule
BERT & surprisal+ROI & 66.10 & 1.00 & $<.001$ & 64.10\\
 & $\Delta$surprisal$\times$ROI & 16.19 & 5.00 & 0.01 & 70.28\\
\midrule
GPT-2 & surprisal+ROI & 98.95 & 1.00 & $<.001$ & 96.95\\
  & $\Delta$surprisal$\times$ROI & 63.90 & 5.00 & $<.001$ & 150.85\\
\bottomrule
\end{tabular} 
}
    \end{minipage}%
    \begin{minipage}{.5\textwidth}
        \centering
        \caption*{Model comparisons in audio-visual setting}
        \label{tab:second}
        \resizebox{0.8\linewidth}{!}{
\begin{tabular}{lrrlrr} 
\toprule
& Model & $\tilde{\chi}^2$ & Df & $p$ & $\tilde{\Delta}AIC$ \\
\midrule
$2$-gram & surprisal+ROI & 246.45 & 1.00 & $<.001$ & 244.45 \\
 & $\Delta$surprisal$\times$ROI & 68.26 & 5.00 & $<.001$ & 302.71 \\
\midrule
$3$-gram & surprisal+ROI & 153.54 & 1.00 & $<.001$ & 151.54 \\
 & $\Delta$surprisal$\times$ROI & 91.27 & 5.00 & $<.001$ & 232.82 \\
\midrule
$4$-gram & surprisal+ROI & 129.53 & 1.00 & $<.001$ & 127.53 \\
 & $\Delta$surprisal$\times$ROI & 100.59 & 5.00 & $<.001$ & 218.12 \\
\midrule
$5$-gram & surprisal+ROI & 129.72 & 1.00 & $<.001$ & 127.72 \\
 & $\Delta$surprisal$\times$ROI & 101.96 & 5.00 & $<.001$ & 219.69 \\
\midrule
$6$-gram & surprisal+ROI & 134.70 & 1.00 & $<.001$ & 132.70 \\
 & $\Delta$surprisal$\times$ROI & 101.51 & 5.00 & $<.001$ & 224.21 \\
\midrule
BERT & surprisal+ROI & 147.00 & 1.00 & $<.001$ & 145.00 \\
 & $\Delta$surprisal$\times$ROI & 27.15 & 5.00 & $<.001$ & 162.15 \\
\midrule
GPT-2 & surprisal+ROI & 19.45 & 1.00 & $<.001$ & 17.45 \\
 & $\Delta$surprisal$\times$ROI & 87.45 & 5.00 & $<.001$ & 94.91 \\
\bottomrule
\end{tabular} 
}
    \end{minipage}
\caption*{Tables $I$: Model comparisons in audio-only (left) and audio-visual (right) settings. $\tilde{\chi}^2$ and $p$-values of additive (surprisal$+$ROI) and multiplicative (surprisal$\times$ROI) models were derived from comparisons with baseline and additive models respectively, while $\tilde{\Delta}AIC$ was always derived from comparisons with the baseline model. We observe that, models with surpirsal (both additive and muliplicative models), in both settings, provide a good fit for N400 amplitudes. We also observe that multiplicative models, almost always, provide a better fit than additive models.}
\end{table*}

\subsection{Statistical Analysis}

\subsubsection{Correlation between Audio and Audiovisual N400}

To determine the correlation between N400 in audio and audio-visual settings, we calculate Pearson's correlation of N400 per word across modalities. N400 was calculated as the mean ERP between 300-500ms minus the baseline ERP mean (as we did not perform baseline correction during preprocessing, as previously mentioned). The variance was reduced by averaging the results across all participants and electrode sites for each word in each modality.

\subsubsection{Evaluating model performances across modalities}

We compared the performances of surprisal generated by different computational models using a linear mixed effect regression model conducted in R using the lme4 package \citep{bates2010lme4}. We followed a similar approach as 
\citet{michaelov2021different} by comparing a baseline model with more complex models containing surprisal. The dependent variable was the mean ERPs in the 300-500ms time window extracted from 32 electrodes for all content words (e.g. nouns, verbs, adjectives, as in \citet{frank2015erp}). The baseline model contains regions of interest (ROI) which describes the location of each electrode. 32 electrodes were catogorised as 5 ROIs, including prefrontal (Fp1, Fp2, AF3, AF4), fronto-central (F3, F7, Fz, F4, F8, FC5, FC1, FC6, FC2), central (C3, C4, Cz), posterior (CP1, CP5, CP2, CP6, P3, P7, Pz, P4, P8, PO3, PO4, O1, Oz, O2), left temporal (T7) and right temporal (T8). The baseline model also contains the mean EEG amplitude from the baseline interval extracted above. The baseline model includes participant, passage and electrode as random intercepts. Then, we added the main effect of surprisal to create surprisal+ROI models and further the interaction between surprisal and ROI to create surprisal$\times$ROI models. We then estimated the improvement of fit by model comparisons, where the surprisal+ROI models are compared with the baseline models and the surprisal$\times$ROI models are compared with the surprisal+ROI models using \texttt{anova()} function in R ($p$-values FDR adjusted for multiple comparisons). We also calculated the decrease of AIC value for each model compared with the baseline model ($\tilde{\Delta}AIC$). The same analysis was performed for audio and audio-visual data separately.

 \begin{figure*}[!h]
  \begin{center}
  \includegraphics[width=0.9\linewidth]{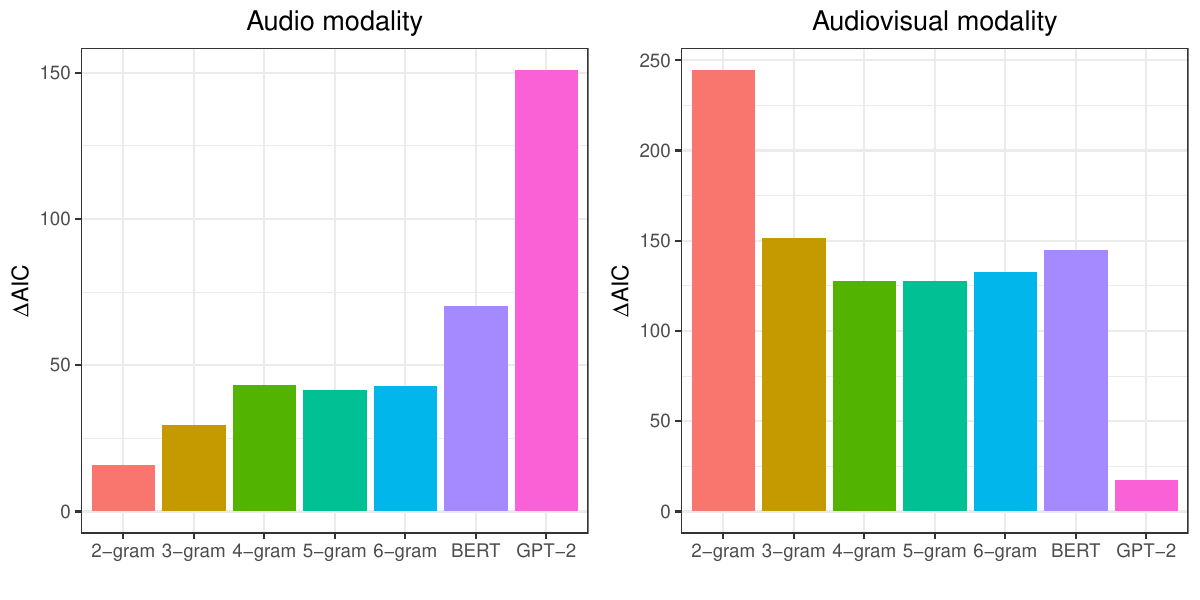}
  \caption{Reduction of AIC (in surprisal$\times$ROI models) associated with each model across modalities. While models with access to larger lexical context windows provide a better fit in the audio-only setting, models with smaller and local lexical context seem to provide a better fit in the audio-visual setting for same verbal stimuli.} 
  \label{figure 2}
  \end{center}
\end{figure*}
\section{Results}

\subsection{N400 is weakly correlated across settings}

The Pearson correlation coefficient for N400 per word between audio-only and audio-visual settings is 0.11 (t = 5.16, $p<.001$), indicating a weak positive correlation between the two settings. %
Figure \ref{fig1} shows a scatter plot of N400 across the two settings, which indicates that while most of the data points are densely populated in the center, there is no meaningful relationship between the two settings. If the lexical information were the most significant contributing factor, we would expect a stronger correlation between audio-only and audio-visual conditions since both experiments involve the same verbal stimuli.

\subsection{Statistical models behave differently across settings}

We present statistical analysis for the audio-only and audio-visual settings in Tables $I$. We find that additive models (surprisal$+$ROI) provide a good fit for N400 amplitudes than baseline models in both audio-only and audio-visual settings, as indicated in $\tilde{\chi}^2$ and $p$ values. Furthermore, the multiplicative models (surprisal$\times$ROI) almost always improve the model fit compared to the additive models. The difference of the multiplicative models over additive models (surprisal$\times$ROI) indicates that surprisal generated from all models predicts N400 amplitudes in both audio and audio-visual conditions in interaction with ROIs.

In the auditory setting, we observe the largest reduction in AIC compared to the baseline model ($\tilde{\Delta}AIC$) is associated with GPT-2, followed by BERT and n-gram models. This suggests that Transformer-based models (especially GPT-2), which has access to a largest lexical context, can better predict N400 amplitudes, in a unimodal setting. Previous work has also seen similar pattern, where models that consider larger lexical contexts have been shown to provide better fit \citep{michaelov2021different,meister2021revisiting,frank2015erp}.

However, in the audio-visual setting, we observe a reversal of this pattern where, strikingly, the 2-gram model shows the largest $\Delta$AIC, while GPT-2 shows the smallest $\Delta$AIC. In general, we notice that the models with smaller context window provide a better fit in the audio-visual setting. We present our results in Figure 2, which shows the reduction in AIC ($\Delta$AIC) across models and modalities. We note that we only plot multiplicative models, as they offer a better overall fit (but the additive models showed similar patterns). These observations indicate that local lexical information is more prominent in the multimodal setting.

\section{Discussion}

Our results demonstrate that under the same verbal stimuli, cognitive processing, as captured using ERP, significantly differs between unimodal and multimodal experimental settings. 
We replicate the earlier findings of multiplicative models (surpirsal$\times$ROI) providing better fit for the data in comparison to additive models (surpirsal$+$ROI) models. Although we validate earlier findings that Transformer-based models like GPT-2 are better predictors of N400 in unimodal (audio-only) settings, the opposite trend is observed in the multimodal setting. These observations strongly suggest that non-verbal cues significantly contribute to cognitive processing more than lexical information alone.

In the unimodal setting, the surprisal estimates from GPT-2 based language model exhibit the best fit compared to other models, consistent with previous research \citep{michaelov2022more} demonstrating the superiority of Transformer-based models over other language models, such as RNNs and traditional $n$-gram models over a variety of psychometric data. Our findings show that in the unimodal setting, the surprisal estimates from GPT-2 based language model outperforms other models, as previously demonstrated in previous studies. However, the BERT displays slightly different results, possibly due to its training objective as a masked language model, which limits access to only pseudo log-probabilities. This difference in objectives between BERT and GPT-2, combined with the limitations in accessing log-probabilities from BERT, could contribute to the differing performance of these models. Similar findings have been reported in previous work \citep{meister2021revisiting}.

In the multimodal setting, our results reveal a reversal of trends compared to the unimodal setting. Surprisal values derived the $n$-gram language models, particularly the $2$-gram model, provide the best fit for N400 in the multimodal scenario. We note that surprisal only captures word predictability based on previous lexical context, ignoring any multimodal information in the stimuli. We posit that, in the multimodal setting, participants utilise multiple sources of information, such as gestures, mouth movements, eye movements, and posture. The increased information content from multiple sources may only allow participants to better track local lexical context, rather than global lexical context. Our findings using language models over different contextual windows suggest some validation of this hypothesis. Especially, we observe in Figure 2 that as we increase the context window from 2 to 6 we overall see a degradation in $\Delta$AIC, indicating a worse fit in comparison to 2-gram. The differences in N400 across audio and audiovisual modalities indicate that cognitive processing strategies differ across modalities even when the verbal stimuli is identical. 

Overall, our findings provide strong evidence that multimodal processing of language differs significantly from unimodal processing of language, even under the same verbal stimuli. Our results generally highlight the importance of considering non-verbal cues in language processing.

\section{Summary and Conclusions}
In this paper, we present a controlled study, investigating the effect of multiple modalities of information on cognitive processing of language comprehension. We conduct experiments over audio-only and audio-visual modalities with the same verbal stimuli. 
Our findings overall suggest that cognitive effort in a multimodal setting significantly differs from that in a unimodal setting, with nonverbal contextual information playing a significant role. We also observe that local verbal context significantly influences cognitive processing effort in a multimodal setting in comparison to the unimodal setting. We believe that our results highlight the importance of modelling non-verbal cues for language comprehension and processing.\footnote{Sourcecode and data for replication of our study are made available here: \url{https://github.com/pmadhyastha/multimodal_comprehension}}

\printbibliography

\end{document}